\pdfoutput=1
\documentclass[letterpaper,10pt,conference]{ieeeconf}

\IEEEoverridecommandlockouts
\usepackage[table]{xcolor}
\usepackage{cite}
\usepackage{amsmath}
\usepackage{amssymb}
\usepackage{booktabs}
\usepackage{graphicx}
\usepackage{grffile}
\usepackage{subcaption}
\usepackage{algorithm}
\usepackage{algpseudocode}
\usepackage{tabularx}
\usepackage{cuted}
\usepackage{etoolbox}
\apptocmd{\thebibliography}{\interlinepenalty=10000\relax\itemsep=0pt\relax}{}{}

\newcommand{\tabnote}[1]{\par\vspace{3pt}{\footnotesize\raggedright\textit{Note:} #1\par}}
\usepackage[listings]{tcolorbox}
\usepackage{xspace}
\usepackage{placeins}
\usepackage{pifont}

\definecolor{Gray}{gray}{0.9}
\definecolor{White}{gray}{1}
\definecolor{DGray}{gray}{0.8}
\definecolor{DDDDGray}{gray}{0.3}
\definecolor{WhiteGray}{rgb}{0.9, 0.9, 0.9}
\definecolor{citecolor}{HTML}{0071bc}
\definecolor{darkred}{rgb}{0.6, 0.1, 0.05}
\definecolor{DeltaColor}{rgb}{0.039,0.73,0.71}
\definecolor{SigmaColor}{rgb}{0.98,0.45,0.0}
\definecolor{AlphaColor}{rgb}{0,0,0.8}
\definecolor{BetaColor}{rgb}{0.8,0,0.8}
\definecolor{GammaColor}{rgb}{0.514,0.34,0.224}
\definecolor{EpsilonColor}{rgb}{0.353,0.725,0.906}
\definecolor{PurpleColor}{HTML}{8B008B}
\definecolor{BadColor}{HTML}{C0392B}
\definecolor{OrangeColor}{rgb}{0.914,0.541,0.0.141}
\definecolor{GreenColor}{rgb}{0.137,0.573,0.565}
\definecolor{RedColor}{rgb}{0.949,0.275, 0.224}
\definecolor{LightCyan}{rgb}{0.88,1,1}
\definecolor{Gray}{gray}{0.85}

\definecolor{bestcolor}{rgb}{1, 0.5, 0.25}
\definecolor{secondbestcolor}{rgb}{1, 0.8, 0.5}

\newcommand{\del}[1]{}

\DeclareMathAlphabet\mathbfcal{OMS}{cmsy}{b}{n}

\newlength\savewidth

\newcommand{\method}{\textbf{FORGE}\xspace}

\newcommand{\cls}{\texttt{CLS}\xspace}

\usepackage[hidelinks]{hyperref}

\title{\LARGE \bf
\method: Force-Aware Multi-Granularity Control for Efficient Humanoid Loco-Manipulation
}

\author{%
Liangjun Xing$^{1,2,*}$, Ke Fan$^{1,4,*}$, Kailin Li$^{1,*}$, Bin Wang$^{1,3}$, Xiaojie Niu$^{1}$,\\
Qingwei Ben$^{1,5}$, Jiafei Cao$^{1}$, Wei Yao$^{2}$, Yebin Liu$^{2,\dagger}$, Jingbo Wang$^{1,\dagger}$, Weinan Zhang$^{1,4,\dagger}$\\[3pt]
{\small $^{1}$Shanghai AI Laboratory\quad $^{2}$Tsinghua University\quad $^{3}$Zhejiang University}\\
{\small $^{4}$Shanghai Jiao Tong University\quad $^{5}$The Chinese University of Hong Kong}\\[2pt]
{\small $^{*}$Equal contribution.\quad $^{\dagger}$Corresponding authors.}%
}

\begin{document}
\maketitle
\thispagestyle{empty}
\pagestyle{empty}

\begin{strip}
\par\vspace{-36pt}
\centering
\includegraphics[width=0.90\textwidth]{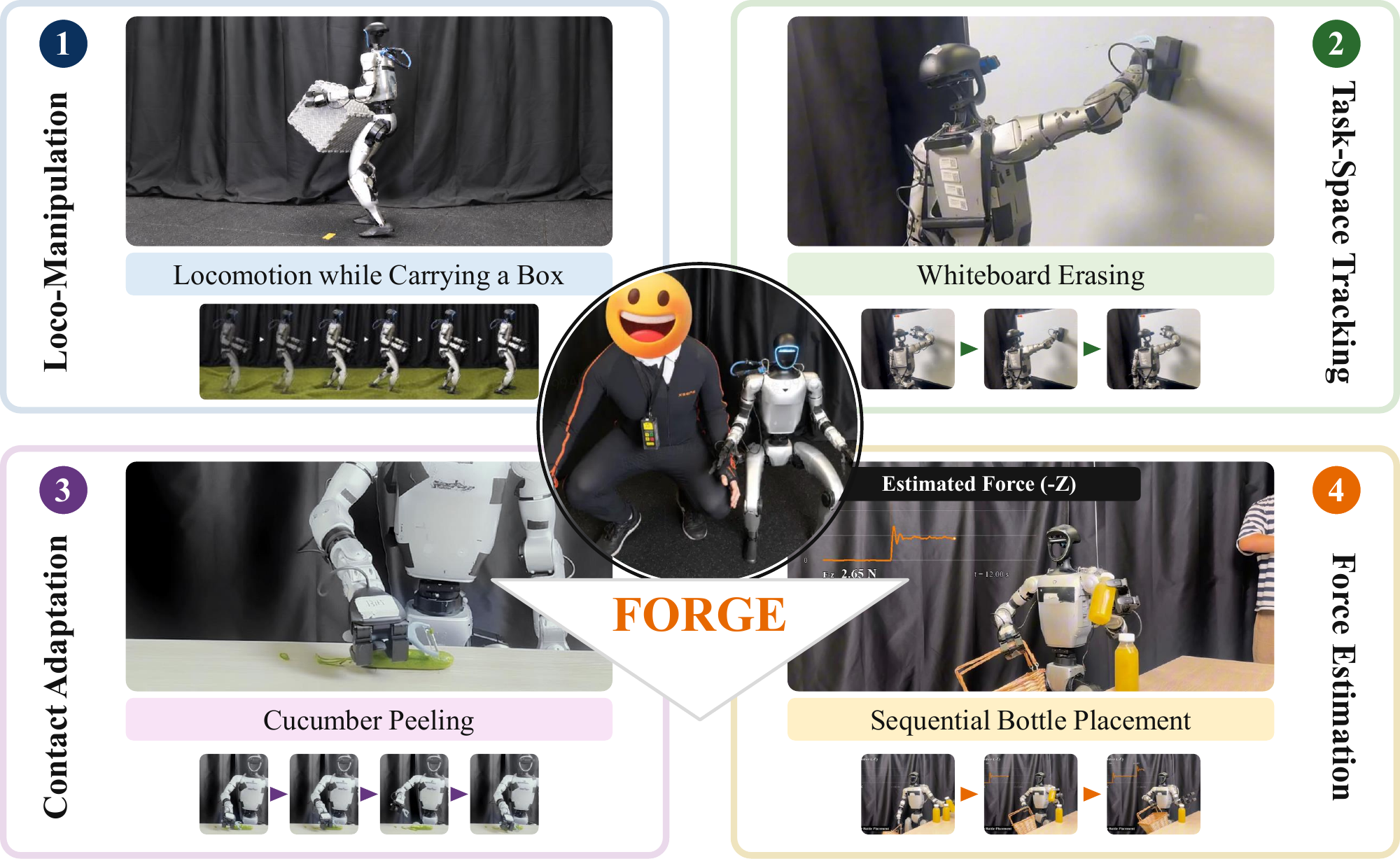}
\captionof{figure}{\method enables efficient humanoid loco-manipulation via multi-granularity control: velocity navigation, foot placement, single-/dual-arm end-effector tracking, and force-conditioned contact adaptation.}
\label{fig:front_cover}
\end{strip}

\begin{abstract}
Humanoid loco-manipulation requires coordinated locomotion, balance regulation, and manipulation under frequent contacts, making teleoperated demonstration collection difficult: a practical controller must simultaneously provide precision, stability, and contact awareness. We propose \method, a force-aware multi-granularity control system with a coarse-to-fine command space spanning velocity navigation, planar foot placement, wrist-pose tracking, and optional force-conditioning inputs. Policy training requires no full-body human motion trajectories or joint-angle references; deployment consumes only sparse bilateral wrist poses relative to the pelvis, coarse motion commands, and optional desired wrist-load inputs. For sustained contact, \method combines a momentum observer with a learned correction trained using simulator contact-force supervision. At deployment, it estimates low-frequency, wrist-equivalent interaction forces from proprioception and reconstructed low-level arm PD torques, without wrist force/torque sensors. A quasi-static virtual-target loop uses this estimate and an optional desired applied-force input to adapt the wrist target. A frozen locomotion prior and asymmetric residual actions stabilize manipulation-induced whole-body corrections. Experiments evaluate unified-control stability, task-space tracking, quasi-static wrist-equivalent force estimation, squeezing forces during sustained contact with a rigid box, and downstream diffusion-policy execution of a squat-and-lift task across payload conditions. Website: \href{https://bb0928.github.io/forge/}{\texttt{bb0928.github.io/forge}}.
\end{abstract}

\section{Introduction}
Humanoid loco-manipulation requires coordinating manipulation with stepping, leaning, or posture adjustment under complex dynamics and frequent contacts \cite{he2025asap,ben2025gallant,xu2025facetforceadaptivecontrolimpedance}. Unlike fixed-base dual-arm robots, which benefit from stable support and mature teleoperation systems \cite{fu2024mobile}, humanoids must map operator intent to whole-body behavior under underactuation and changing contacts. Balance errors can compromise stability, joint errors can accumulate into end-effector drift, and rigid position control can disrupt sustained interaction. A practical controller must therefore jointly provide \emph{Stability}, \emph{Precision}, and \emph{Contact Awareness}, including interaction-force estimation for regulating sustained contact.

Existing approaches satisfy these requirements only in part. Full-body tracking and large-scale motor-control models \cite{he2024omnih2o,luo2025sonic,liao2025beyondmimicmotiontrackingversatile} provide broad motion repertoires, but dense motion-reference objectives do not directly expose task-space commands for precise foot placement, wrist targeting, and sustained contact. Decoupled locomotion/upper-body stacks \cite{DBLP:journals/corr/abs-2502-13013,DBLP:journals/corr/abs-2506-08931} preserve locomotion stability but can limit task-dependent whole-body coordination. Force-aware approaches \cite{unifp,lu2025gentlehumanoidlearningupperbodycompliance,wang2025physhsirealworldgeneralizablenatural} demonstrate learned force estimation and force-conditioned position control, yet sensorless estimation is difficult: external contact, friction, actuator behavior, and whole-body coupling are entangled in proprioception.

A unified controller must therefore support coarse-to-fine teleoperation: approaching a task region, adjusting stance, tracking wrist trajectories, and adapting to contact. Commands must remain composable as task demands change.

These limitations motivate \method, which \emph{forges} a \textbf{FOR}ce-aware multi-\textbf{G}ranularity policy for \textbf{E}fficient humanoid loco-manipulation. Its unified coarse-to-fine interface supports base velocity commands, planar foot-placement targets, Cartesian wrist-pose targets, and optional force-conditioning inputs, individually or jointly. Granularity-conditioned masking \cite{DBLP:journals/tog/TesslerGNCP24} activates the observations and rewards relevant to each command group. Training needs no full-body human trajectories or human joint-angle references; teleoperation supplies only bilateral wrist poses relative to the pelvis and coarse commands from wearable trackers, handheld VR devices, or vision-based pose estimation.


For sustained contact, a \emph{Physics-Informed Hybrid Force Estimator} learns a simulator-supervised correction to a model-based momentum observer. At deployment, it uses arm joint states and reconstructed low-level PD torques without wrist force/torque sensors. The estimator captures the low-frequency component of the equivalent interaction force at each wrist. We combine this estimate with an optional command for the robot-applied force to adjust wrist position targets during sustained, quasi-static contact.

To stabilize mixed-objective training, we freeze a pretrained lower-body locomotion expert and learn an asymmetric residual policy that controls the upper body and applies corrective lower-body actions. This reduces exploration difficulty while preserving the locomotion prior and allowing task-dependent whole-body coordination.



Simulation and hardware experiments show that masking and asymmetric residual training improve unified-control stability and task-space tracking over their ablations, while the hybrid estimator reduces force-estimation error under the evaluated quasi-static contacts compared with analytical-only and learned-only estimators. To assess \method as a whole-body executor for visuomotor learning, we train a diffusion policy \cite{dp} on collected demonstrations; it commands \method to execute squat-and-lift across payloads.

Our contributions are a unified multi-granularity controller trained without full-body human motion references, physics-informed sensorless wrist-force estimation with contact-load adaptation, and asymmetric residual learning for stable whole-body loco-manipulation.

\begin{figure*}[t]
    \centering
    \includegraphics[width=0.90\linewidth]{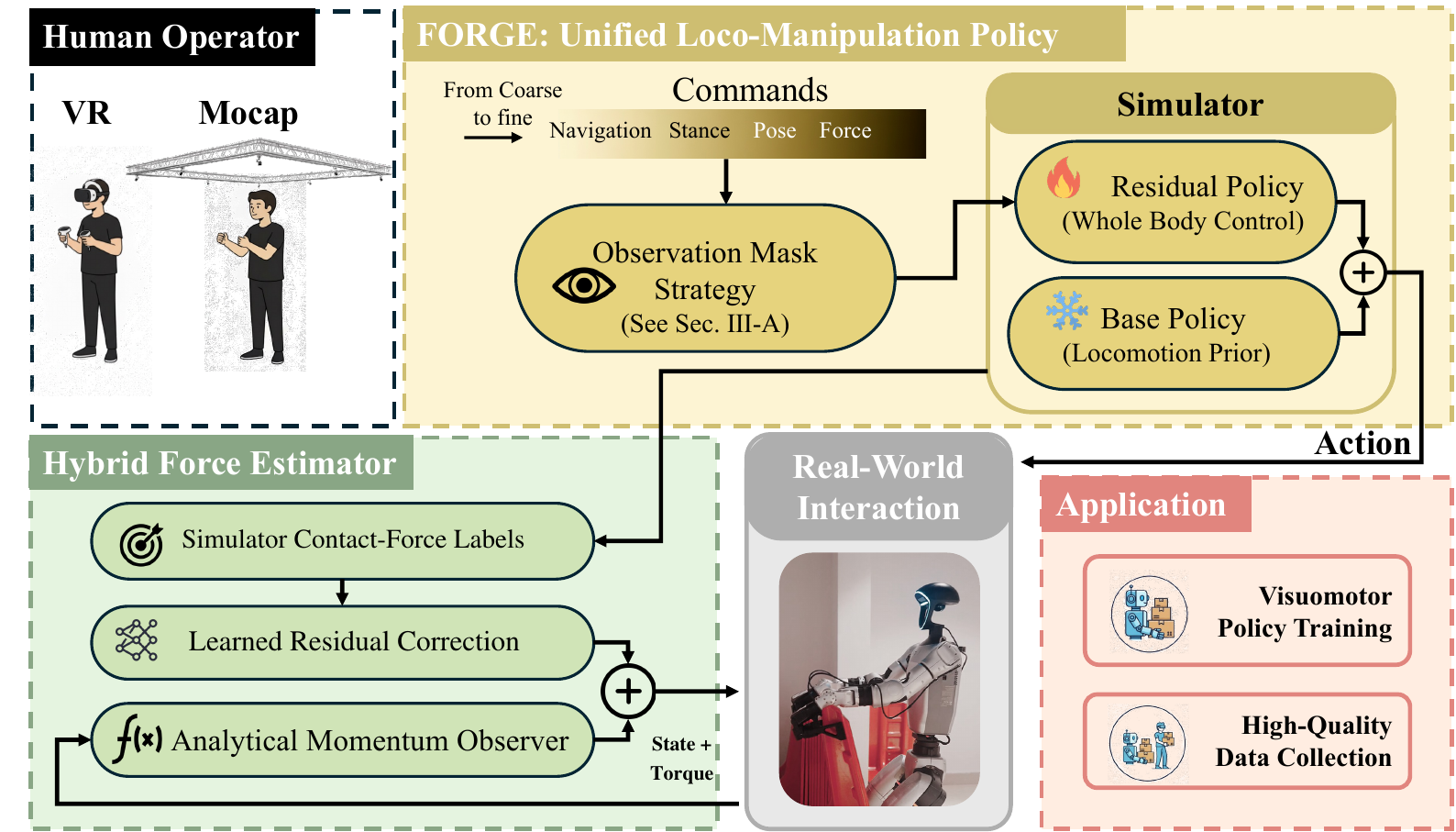}
    \caption{\textbf{Overall pipeline of \method.} The framework has three components: multi-granularity control over navigation, stance, pose, and optional force-conditioning inputs; a physics-informed hybrid force estimator and virtual-target contact-adaptation loop; and asymmetric residual actions anchored on a frozen locomotion prior. Policy training uses sparse task-space commands rather than full-body human motion references.}
    \label{fig:forge_pipeline}
\end{figure*}

\section{Related Work}

\subsection{Whole-Body Control and Teleoperation Systems}
\label{sec:rw_architecture}

Humanoid loco-manipulation controllers broadly follow two paradigms. Motion-reference-driven policies learn broad whole-body behaviors from retargeted or captured motions \cite{he2024learning,he2024omnih2o,liao2025beyondmimicmotiontrackingversatile,luo2025sonic,ze2025twist2}, whereas modular systems separate learned locomotion from upper-body control \cite{DBLP:journals/corr/abs-2502-13013,DBLP:journals/corr/abs-2506-08931}. The former relies on dense whole-body references; the latter preserves locomotion stability but can restrict task-dependent whole-body coordination. SoFTA explicitly separates slow locomotion and fast upper-body control to stabilize the end effectors during walking \cite{li2025holdbeerlearninggentle}. Related work also explores masked motion generation and reusable behavior priors \cite{DBLP:journals/tog/TesslerGNCP24,tessler2025maskedmanipulatorversatilewholebodymanipulation}. In contrast, \method trains from task-space commands without full-body human motion references and uses a frozen locomotion prior with asymmetric residual learning \cite{rudin2022learning,li2025maniptrans}. Its teleoperation input comprises only pelvis-relative wrist poses and coarse commands, whether obtained from wearable, handheld, or vision-based trackers.

Building on 3D Diffusion Policy~\cite{dp3}, Improved 3D Diffusion Policy combines upper-body teleoperation with 3D visuomotor learning on a cart-supported humanoid \cite{idp3}. Our downstream evaluation instead uses a diffusion policy to command a balancing whole-body executor.

\subsection{Force Perception and Compliant Interaction}
\label{sec:rw_force}

Humanoid controllers increasingly incorporate interaction-aware or force--position objectives \cite{zhang2025falcon,chen2025chipadaptivecompliancehumanoid}, while sensorless estimators infer wrist interaction from proprioception \cite{unifp,lu2025gentlehumanoidlearningupperbodycompliance,wang2025physhsirealworldgeneralizablenatural}. UniFP estimates external force from proprioceptive history and maps force--position commands to compensated position targets \cite{unifp}. \method retains this separation but adds a momentum-observer prior before learned correction. It estimates low-frequency translational wrist interaction for quasi-static virtual-target adaptation, not full-wrench sensing or dynamic impedance control.


\section{Method}

\method combines a compact multi-granularity command interface with a force-aware contact pathway. Navigation, stance, and wrist-pose commands share one whole-body policy; a \textbf{Physics-Informed Hybrid Force Estimator} and virtual-target loop provide sustained-contact adaptation without wrist force/torque sensors. \textbf{Asymmetric Residual Actions} anchor this policy on a frozen locomotion prior (Fig.~\ref{fig:forge_pipeline}).

\subsection{Multi-Granularity Control System}
\label{sec:multi_granularity}

One policy accepts four individually or jointly active command groups: base velocity $\mathbf v_{\mathrm{cmd}}$ (\textbf{Navigation}), planar left/right footholds $\mathbf p_{\mathrm{feet}}\in\mathbb R^{2\times2}$ (\textbf{Stance}), bilateral wrist poses $\mathbf T_{\mathrm{wrist}}\in\mathrm{SE}(3)^2$ (\textbf{Pose}), and optional desired robot-applied wrist forces $\mathbf f_{\mathrm{cmd}}\in\mathbb R^{2\times3}$ (\textbf{Force}). An operator or upstream model may supply the force command for virtual-target updates (Sec.~\ref{sec:hybrid_force_estimator}), not direct torque control.

During training, we sample active combinations of these four command groups and their target values. For each inactive group, we replace the associated observation entries with null values and exclude its reward terms from the training objective. Masking both observations and objectives prevents stale targets from conflicting with the active task. Four task-level activation bits expand into a per-group mask supplied with a six-step observation history; compound behaviors activate the required groups without selecting another controller or changing the 29-D action semantics. The deployed one-step signal contains 138 scalars: 3 navigation, 5 stance, 17 pose, and 6 desired-force command entries, together with 6 IMU quantities, 29 joint positions, 29 joint velocities, the previous 29-D action, and the 14-D momentum-observer residual. The explicit semantic-group mask is appended separately and excluded from this count.

\begin{figure}[t]
    \centering
    \includegraphics[width=\linewidth]{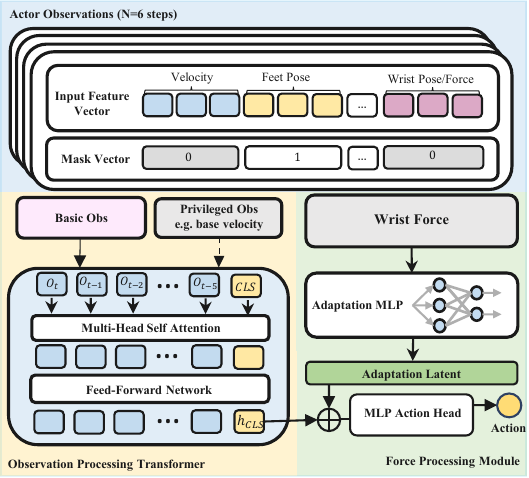}
    \caption{Actor architecture. Solid deployment paths use only the masked observable history. ``Wrist Force'' denotes the observer-derived residual history, not a force/torque-sensor reading or simulator force. The adaptation decoder's 6-D correction refines the observer at deployment; the dashed privileged quantities supply training targets only.}
    \label{fig:transformer_policy}
\end{figure}

As shown in Fig.~\ref{fig:transformer_policy}, every step is partitioned into semantic groups for base motion, footholds, wrist poses, desired wrist forces, the 14-D observer residual, and proprioception. After the mask is applied, one shared MLP maps each per-step vector to a token. A two-layer Transformer (128-D latent, four heads, 256-D feed-forward layers) uses a learnable \cls token to summarize the sequence as $h_{\cls}$. In parallel, a 512--256--128 adaptation encoder maps the same six masked observable frames to a 64-D latent $z_{\mathrm{adapt}}$; the 512--256--128 action head consumes $[h_{\cls},z_{\mathrm{adapt}}]$. A 128--64 decoder maps $z_{\mathrm{adapt}}$ to 15 supervised outputs: a 6-D observer-force correction, bilateral wrist position (6-D), and base linear velocity (3-D). The first six outputs remain active at deployment and refine the observer estimate; wrist position and base velocity are auxiliary training targets. Simulator contact forces supervise the corrected force estimate, while simulator wrist positions and base linear velocities supervise the auxiliary predictions during training; none of these reference values is supplied to the actor.

\begin{figure}[t]
    \centering
    \begin{subfigure}[b]{0.40\linewidth}
        \centering
        \includegraphics[height=2.7cm]{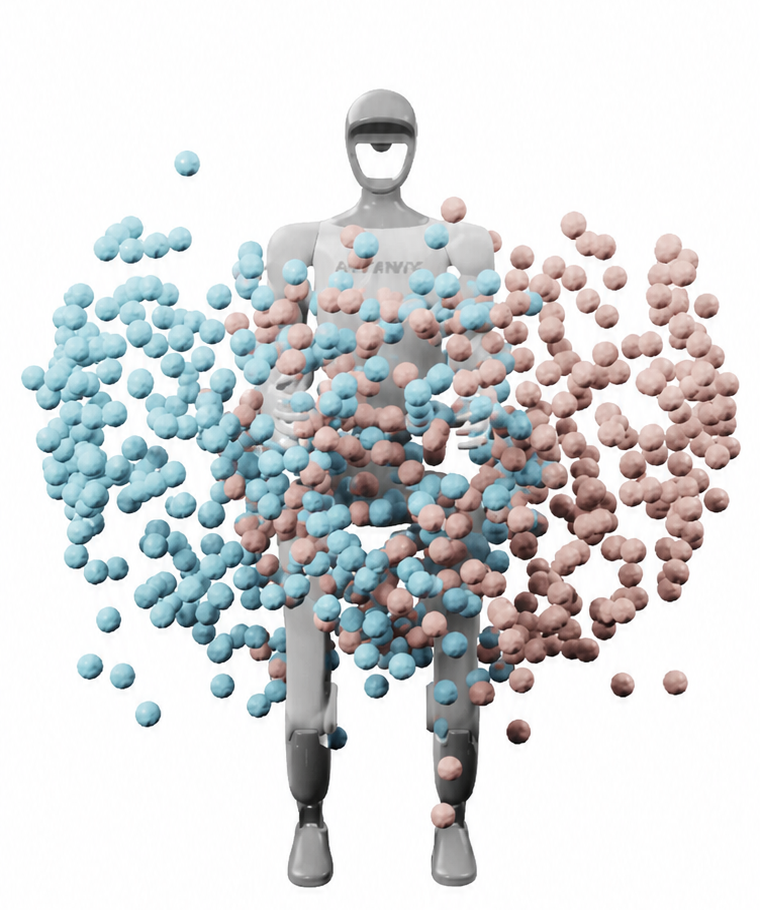}
        \caption{Front view.}
    \end{subfigure}\hspace{0.04\linewidth}
    \begin{subfigure}[b]{0.40\linewidth}
        \centering
        \includegraphics[height=2.7cm]{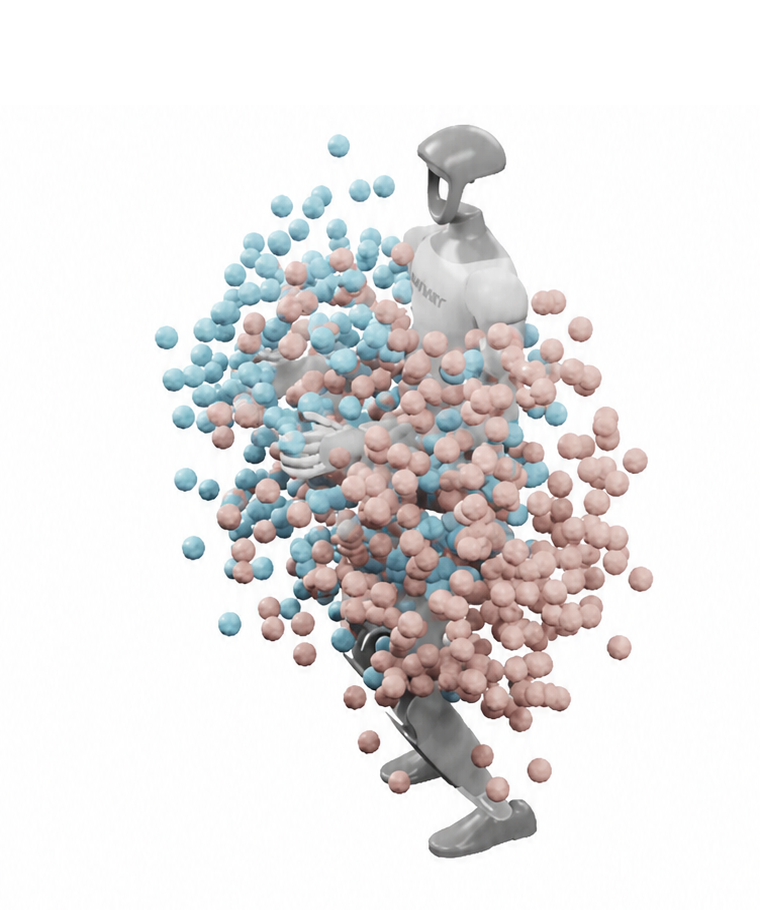}
        \caption{Side view.}
    \end{subfigure}
    \caption{Kinematics-aware bilateral wrist-target distributions.}
    \label{fig:wrist_sampling_dist}
\end{figure}

Stance targets are sampled within a reachable support region and expressed in the base-yaw frame. For Pose targets, we interpolate valid upper-body joint anchors, map them through wrist forward kinematics, and rotate the resulting Cartesian targets into the base-yaw frame. This procedure produces the reachable distributions in Fig.~\ref{fig:wrist_sampling_dist}; the actor observes wrist targets rather than the sampled joints and resolves whole-body IK implicitly.

\subsection{Physics-Informed Hybrid Force Estimator}
\label{sec:hybrid_force_estimator}

While the Multi-Granularity Control System specifies targets, the \emph{Force} granularity requires contact feedback. For each wrist $i\in\{L,R\}$, we distinguish the optional desired robot-applied force $\mathbf f_{\mathrm{cmd}}^i$, the wrist-equivalent environmental reaction $\mathbf f_{\mathrm{env}}^i$, and its estimate $\hat{\mathbf f}_{\mathrm{env}}^i$. The desired and environmental forces use the same world-frame axes but represent opposite sides of the contact pair; consequently, ideal quasi-static force balance satisfies $\mathbf f_{\mathrm{env}}^i=-\mathbf f_{\mathrm{cmd}}^i$. Here ``wrist-equivalent'' denotes the 3D translational force whose Jacobian transpose best explains the observed arm-joint disturbance under the assumed wrist-contact condition; unmodeled link contacts and whole-body coupling are not separately identified. Rather than using dedicated wrist force/torque sensors, we estimate its low-frequency, quasi-static component from arm joint states and reconstructed low-level PD torques. The estimator combines a \emph{model-based momentum observer}~\cite{deluca2005sensorless} with a \emph{learned correction}: the observer supplies a physically structured disturbance estimate, while the learned term compensates repeatable residual errors from the reduced arm model and actuation.

\subsubsection{Model-Based Momentum Observer}

We first estimate a joint-space disturbance residual using an arm momentum observer~\cite{deluca2005sensorless}, avoiding direct acceleration differentiation. Let $\mathbf q^u\in\mathbb R^{14}$ contain the two 7-DoF arm configurations and let $\mathbf M^u(\mathbf q^u)$ assemble their arm-only mass matrices. With internal momentum state $\hat{\mathbf p}$ and observer residual $\mathbf c\in\mathbb R^{14}$, each control step applies
\begin{align}
\hat{\mathbf p}_{k+1}
&=\hat{\mathbf p}_{k}+\Delta t\big(\boldsymbol{\tau}_m-\hat{\mathbf g}-\hat{\boldsymbol{\tau}}_f+\mathbf c_k\big),\\
\mathbf c_{k+1}
&=\beta\mathbf c_k+\Delta t\,\mathbf K\big(\mathbf M^u(\mathbf q^u)\dot{\mathbf q}^u-\hat{\mathbf p}_{k+1}\big).
\end{align}
where $\boldsymbol{\tau}_m$ is a PD-torque proxy reconstructed from commanded joint targets and measured joint states, rather than motor-current torque; $\hat{\mathbf g}$ and $\hat{\boldsymbol{\tau}}_f$ model gravity and smooth Coulomb friction. We set $\beta=0.9$ and initialize $\hat{\mathbf p}_0=\mathbf c_0=\mathbf 0$. The assembled arm-only mass matrices, computed with the composite-rigid-body algorithm (CRBA)~\cite{featherstone2008rigid}, preserve within-arm configuration coupling but omit the floating base, waist, legs, and cross-arm coordinates. Consequently, velocity-dependent terms, excluded whole-body coupling, and actuation mismatch are not canceled analytically; they remain in the residual learned by the correction pathway. We split $\mathbf c=[\mathbf c^{L};\mathbf c^{R}]$ and project each 7-D arm residual through its translational wrist Jacobian $\mathbf J_v^i$:
\begin{equation}
\hat{\mathbf f}_{\mathrm{obs}}^i=
\left(\mathbf J_v^i\mathbf J_v^{i\top}+\lambda\mathbf I_3\right)^{-1}
\mathbf J_v^i\mathbf c^i,\qquad i\in\{L,R\},
\end{equation}
where $\lambda$ improves numerical stability near singularities. We use diagonal gain $\mathbf K=20\mathbf I$, smooth Coulomb coefficient $1.5\,$N\,m with $\epsilon=0.05\,$rad/s, joint-residual clipping at 20, and $\lambda=10^{-3}$. The wrist estimate is norm-clipped at $50\,$N and exponentially smoothed with coefficient 0.8; observer update and projection run at the 50\,Hz policy rate.

\subsubsection{Learned Compensation with Privileged Supervision}

The force-correction output of the adaptation decoder is denoted $\Delta\mathbf f_\theta(\mathbf o_{t-5:t})$. Rather than relearning the complete force mapping, it predicts only the discrepancy between the structured observer output and the simulator contact-force reference from the same six masked frames. The final estimate and training objective are
\begin{equation}
\hat{\mathbf f}_{\mathrm{env}}=\hat{\mathbf f}_{\text{obs}}+\Delta \mathbf f_\theta(\mathbf o_{t-5:t}),\quad
\mathcal{L}_{\text{est}}=\mathbb{E}\left[\left\|\hat{\mathbf f}_{\mathrm{env}}-\mathbf f_{\mathrm{sim}}\right\|_2^2\right].
\end{equation}
Simulator contact forces $\mathbf f_{\mathrm{sim}}$ supervise the corrected estimate and never enter the actor. At deployment, the decoder predicts corrections from observable history; $\mathbf f_{\mathrm{cmd}}$ is the target-adaptation command, not the force-estimation target.

\subsubsection{Force-Conditioned Virtual-Target Adaptation}

Following the force--position separation used in UniFP~\cite{unifp}, we use the desired applied-force input to bias a position target. For each wrist $i\in\{L,R\}$, the desired environmental reaction is $\mathbf f_{\mathrm{env}}^{i,*}=-\mathbf f_{\mathrm{cmd}}^i$. Around the nominal wrist target $\mathbf p_w^{i,*}$, we define
\begin{equation}
\mathbf e_f^i=\hat{\mathbf f}_{\mathrm{env}}^i-\mathbf f_{\mathrm{env}}^{i,*}
=\hat{\mathbf f}_{\mathrm{env}}^i+\mathbf f_{\mathrm{cmd}}^i,\qquad
\mathbf p_w^{i,\mathrm{virt}}=\mathbf p_w^{i,*}+\mathbf K_f^{-1}\mathbf e_f^i,
\label{eq:force_virtual_target}
\end{equation}
where the signed force-balance error vanishes when the desired robot-applied force is balanced by the estimated environmental reaction. The diagonal matrix $\mathbf K_f$ has units N/m, and $\mathbf K_f^{-1}$ maps the mismatch to a position offset in m/N. The two wrist updates are computed independently and then supplied jointly as bilateral pose commands to the same whole-body policy. Force estimation and target updates operate in world coordinates; policy-facing commands are rotated into the base-yaw frame. The $50\,$N norm bound and exponential smoothing are applied to the estimated force before its conversion into the virtual-target offset; they are not a separate hard clip on $\mathbf p_w^{i,\mathrm{virt}}$. Thus, the deployment loop is closed entirely through observable arm state and commanded targets: the torque proxy drives the observer, the learned correction refines its environmental-force estimate, and the desired applied-force input together with that estimate updates the wrist target tracked by the policy. This is an algebraic, admittance-like target adaptation.

At each 50\,Hz control step, the hybrid force estimate is updated and used to compute the virtual wrist target. During training, each Cartesian component of $\mathbf f_{\mathrm{cmd}}$ is sampled in $[-20,20]\,$N and smoothly ramped, held, and released, while independently generated simulator contact forces supervise correction. We use $K_f=200\,$N/m in force-command mode. Training uses 2,048 parallel environments, 50 simulation steps/PPO update, and up to 20,000 updates.

\subsection{Asymmetric Residual Whole-Body Control}
\label{sec:asymmetric_residual}

Mixed locomotion, pose, and contact objectives can destabilize gait learning because unrestricted whole-body exploration can sacrifice support behavior to reduce short-horizon manipulation errors. We therefore train a lower-body expert with upper-body actions fixed to zero, freeze it, and learn a whole-body residual $\Delta\mathbf a=[\Delta\mathbf a^\ell;\Delta\mathbf a^u]\in\mathbb R^{29}$. Freezing the expert prevents gradients from manipulation and contact objectives from rewriting the nominal gait, while the trainable lower-body residual still permits command-dependent foothold and balance corrections. The residual policy receives the active multi-granularity command and learns these bounded leg corrections alongside direct upper-body actions. Summing the frozen action and residual before executing one whole-body action ensures that stance adjustment and manipulation-induced posture changes remain coupled rather than handled by independent controllers. With 15 lower-body and 14 upper-body dimensions, the executed action is
\begin{equation}
\mathbf a=\left[\mathbf a_{\mathrm{base}}^\ell+\Delta\mathbf a^\ell;\,\Delta\mathbf a^u\right],
\end{equation}
where $\mathbf a_{\mathrm{base}}^\ell$ is the frozen locomotion action. PPO exploration uses action-standard-deviation limits of $0.2$ for lower-body and $1.0$ for upper-body residuals, protecting the gait prior while retaining upper-body exploration.

\begin{figure*}[!t]
    \centering
    \includegraphics[width=0.92\textwidth]{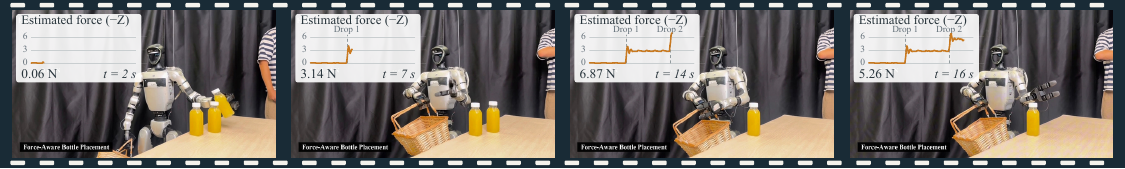}
    \caption{Sequential bottle placement. Four ordered frames show bottles placed into a basket, with vertical wrist-equivalent force traces at marked releases.}
    \label{fig:bottle_force_filmstrip}
\end{figure*}

\section{Experimental Evaluation}

We evaluate four aspects of \method: (i) how its hardware latency and task-space tracking compare with existing humanoid controllers; (ii) whether one policy supports multiple command granularities without destabilizing locomotion; (iii) whether the hybrid estimator improves wrist-equivalent interaction-force estimation and supports contact-load adaptation; and (iv) whether \method serves as a downstream whole-body executor for visuomotor manipulation. Fig.~\ref{fig:bottle_force_filmstrip} shows sequential bottle placement and the estimated vertical wrist-force responses around bottle release.

\subsection{Implementation Details}
Unless otherwise specified, ablation variants share the training settings and environment randomization, except for the components being ablated. Early training fixes yaw-rate commands at zero, smooths torque commands with EMA ($\alpha=0.3$), and applies left--right symmetric augmentation. Task-space termination is disabled for the first 300 steps of every episode and until 2,000 global updates; afterward, we use $\delta_{\text{term}}=0.2\,$m. The lower-body residual action standard deviation is clipped at $0.2$.

The locomotion expert and residual stage use Adam PPO with learning rate $10^{-3}$ under an adaptive schedule targeting KL $0.01$, discount $\gamma=0.99$, GAE parameter $\lambda=0.95$, clip ratio $0.2$, entropy coefficient $0.01$, five learning epochs, four minibatches, and gradient-norm limit $1.0$. Both stages are capped at 20,000 updates. Isaac Gym advances at 200\,Hz with the TGS solver (four position iterations), $0.01\,$m contact offset, and zero rest offset; four-step decimation yields the 50\,Hz policy rate. Actions specify joint targets as $\mathbf q_{\mathrm{des}}=\mathbf q_{\mathrm{default}}+0.25\mathbf a$. The low-level stiffness/damping pairs in N\,m/rad and N\,m\,s/rad are $100/2$ for hips, $150/4$ for knees, $40/2$ for ankles, $300/5$ for the waist, and $40/1$ for shoulder, elbow, and wrist joints.

The actor uses six observable frames, whereas a 512--256--256 critic receives one current privileged frame containing the actor signals plus bilateral simulator wrist force (6-D), wrist position (6-D), virtual target (6-D), and base linear velocity (3-D). Auxiliary losses for wrist force, wrist position, and base velocity use relative weights $1.0/1.0/0.25$, with overall adaptation-loss coefficient $1.0$. Domain randomization samples added base mass in $[-5,10]\,$kg and hand payload in $[-0.1,0.3]\,$kg, link-mass and PD-gain multipliers in $[0.9,1.1]$, COM and body-frame offsets in $[-0.1,0.1]\,$m, friction in $[0.1,2.5]$, and initial-joint scaling/offset in $[0.8,1.2]$/$[-0.1,0.1]\,$rad. External base pushes are applied every 4\,s with velocity perturbations up to $0.5\,$m/s horizontally and $0.2\,$m/s vertically.

\subsection{Comparison with Existing Humanoid Controllers}

Table~\ref{tab:baseline_comparison} compares four controllers on the same robot and target sequences over 20 trials each: \method, HOMIE~\cite{DBLP:journals/corr/abs-2502-13013}, TWIST2~\cite{ze2025twist2}, and SONIC~\cite{luo2025sonic}. For TWIST2 and SONIC, we use the official implementations and checkpoints and reproduce their training pipelines locally. Control latency is the time from teleoperation-command emission to robot end-effector arrival at the target position. Foot and wrist MPE are temporal mean Euclidean target errors in planar 2D and 3D, respectively, averaged over the two limbs and then across trials. HOMIE directly maps exoskeleton commands to upper-body joints. With no wrist-target or foothold interface, tracking is N/A.

We additionally compare the squeezing force generated during sustained bilateral contact in simulation. All methods maintain bilateral contact with the same rigid $0.4\times0.4\times0.4\,$m box, centered at $(0.25,0,0.25)\,$m in the pelvis frame, with the same left/right wrist-position targets at $(0.25,\pm0.15,0.25)\,$m. These targets lie $5\,$cm inside the corresponding box surfaces. From $2\,$s after command onset, we time-average the absolute normal reaction force and average both wrists. This measures each complete controller's squeezing force under identical wrist-position targets.

\begin{table}[!t]
    \centering
    \caption{Comparison with Humanoid Controllers}
    \label{tab:baseline_comparison}
    \footnotesize
    \setlength{\tabcolsep}{2pt}
    \renewcommand{\arraystretch}{1.12}
    \begin{tabularx}{\linewidth}{@{}l*{4}{>{\centering\arraybackslash}X}@{}}
        \toprule
        \textbf{Method} & \textbf{Latency} (ms) $\downarrow$ & \textbf{Foot MPE} (cm) $\downarrow$ & \textbf{Wrist MPE} (cm) $\downarrow$ & \textbf{Contact force} (N) $\downarrow$ \\
        \midrule
        HOMIE  & 27  & N/A  & N/A & 15.48 \\
        TWIST2 & 197 & 7.87 & 7.76 & 17.16 \\
        SONIC  & 252 & 7.35 & 6.42 & 15.32 \\
        \method & \textbf{22} & \textbf{5.42} & \textbf{4.15} & \textbf{8.52} \\
        \bottomrule
    \end{tabularx}
    \tabnote{Latency/tracking: 20 hardware trials per method. Contact force: simulated bilateral mean absolute normal reaction under identical wrist-position targets; lower values indicate less squeezing in this test. Latency: time delay from command transmission to robot end-effector reaching the target position.}
\end{table}

\subsection{Simulation Validation of Unified Control}

The training reward is $R_t=\sum_i w_i r_i$, with all weights scaled by the policy period $\Delta t=0.02\,$s. Tracking terms use $\exp(-\lVert e\rVert^2/\sigma)$ unless noted otherwise, with $\sigma=0.25$ and wrist-position scale $\sigma_w=0.20$. Table~\ref{tab:reward_summary} lists the principal reward terms, weights, and formulations.

\begin{table}[!t]
\centering
\footnotesize
\setlength{\tabcolsep}{2pt}
\renewcommand{\arraystretch}{1.08}
\caption{Principal Reward Terms and Formulations}
\label{tab:reward_summary}
\begin{tabularx}{\linewidth}{@{}>{\raggedright\arraybackslash}p{0.27\linewidth}>{\centering\arraybackslash}p{0.16\linewidth}>{\raggedright\arraybackslash}X@{}}
\toprule
\textbf{Reward Term} & \textbf{Weight} & \textbf{Formulation / Error Term} \\ \midrule
\multicolumn{3}{@{}l}{\textit{\textbf{Command Tracking}}} \\
Vel X / Y Track & 2.0 / 1.5 & $\exp(-\|\Delta v_{x,y}\|^2 / \sigma)$ \\
Ang Vel / Int Track & 2.0 / 2.0 & $\exp(-\|\Delta \omega_z\|^2 / \sigma)$,\newline $\exp(-I_{err} / \sigma_{int})$ \\
Base Height Track & 2.0 & $\exp(-\|\Delta h\| \cdot 10)$ \\ \midrule
\multicolumn{3}{@{}l}{\textit{\textbf{End-Effector Tasks}}} \\
Wrist Pos Track & 5.0 & $\exp(-\|\Delta \mathbf{p}_w\| / \sigma_w)$ \\
Wrist Quat Track & 2.0 & $\exp(-\|2 \text{arccos}(|w_{err}|)\| / \sigma_q)$ \\
Planar Feet Pos Track & 5.0 & $\exp(-\|\Delta \mathbf{p}_{f,xy}\| / \sigma)$ \\ \midrule
\multicolumn{3}{@{}l}{\textit{\textbf{Force-Conditioned Contact Adaptation}}} \\
Virtual-Target Wrist Track & 5.0 & $\exp(-\|\mathbf{p}_w - \mathbf{p}_{virt}\| / \sigma_w)$ \\
Virtual-Target Offset Penalty & 0.01 & $-\text{max}(0, \|\Delta \mathbf{p}_{virt}\| - d_{th})$ \\
Virtual-Target Smoothness & 0.001 & $\exp(-\|\Delta \mathbf{e}_{virt}\| / 0.05)$ \\ \midrule
\multicolumn{3}{@{}l}{\textit{\textbf{Stability \& Posture}}} \\
Orientation & -3.0 & $\|\mathbf{g}_{proj, xy}\|^2$ \\ \midrule
\multicolumn{3}{@{}l}{\textit{\textbf{Penalties}}} \\
Action Rate / Smooth & -0.01 & $\|a_t - a_{t-1}\|^2$ /\newline $\|a_t - 2a_{t-1} + a_{t-2}\|^2$ \\
Feet Slip & -0.25 & $\|v_{f,xy}\|$ \\
\bottomrule
\end{tabularx}
\tabnote{Weights follow the listed terms. Penalties use negative weights, except the offset term with a negative formula.}
\end{table}

\begin{figure*}[t]
    \centering
    \begin{minipage}[t]{0.67\linewidth}
        \centering
        \captionsetup{type=figure}
        \includegraphics[width=\linewidth]{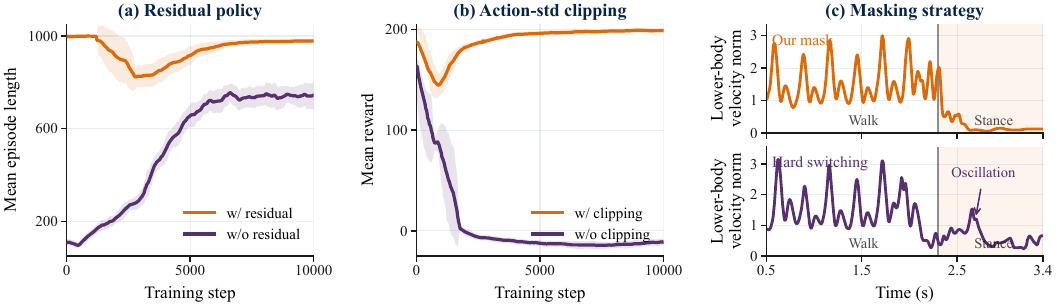}
        \caption{Unified-control ablations. (a) Asymmetric base+residual converges; end-to-end 29-DoF learning does not. (b) Action-std clipping prevents late collapse. (c) Masking smooths walk$\rightarrow$stance transitions.}
        \label{fig:unified_control}
        \label{fig:unified_control_residual}
        \label{fig:unified_control_clip}
        \label{fig:unified_control_switch}
    \end{minipage}\hfill
    \begin{minipage}[t]{0.30\linewidth}
        \centering
        \captionsetup{type=figure,justification=raggedright,singlelinecheck=false}
        \includegraphics[width=\linewidth]{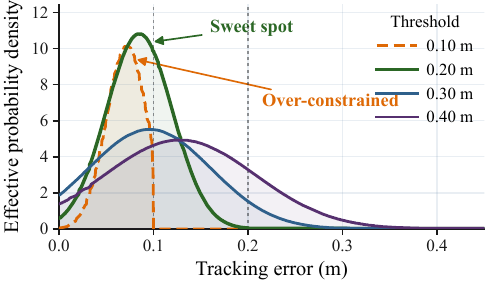}
        \caption{Precision--stability sweep over $\delta_{\text{term}}\in\{0.1,0.2,0.3,0.4\}\,$m.}
        \label{fig:early_term_curve}
    \end{minipage}
\end{figure*}

\noindent\textbf{Unified control, masking, and stability:}
Fig.~\ref{fig:unified_control} and Tab.~\ref{tab:masking_metrics} evaluate whether a single policy can coordinate velocity navigation, foot-placement control, wrist tracking, and force-conditioned contact adaptation while remaining trainable. An end-to-end 29-DoF policy trained without a locomotion prior fails to learn stable behavior, whereas the base--residual design converges. Removing residual action-std clipping causes late-stage reward collapse. The hard-switching baseline in Fig.~\ref{fig:unified_control_switch} corresponds to the ``w/o mask'' ablation in Tab.~\ref{tab:masking_metrics}. For inactive command groups, our method masks the corresponding observations and excludes their task-specific reward terms from training. This ablation increases early-termination frequency and velocity, foot, and wrist errors, consistent with interference between inactive command groups and active objectives. Together, the residual prior, lower-body exploration clipping, and masking support stable multi-granularity training under the evaluated command schedule.

Fig.~\ref{fig:early_term_curve} compares tracking-error distributions under different termination thresholds. The $0.2\,$m setting concentrates errors at low values, whereas $0.3$ and $0.4\,$m yield broader distributions with longer high-error tails; we therefore use $\delta_{\text{term}}=0.2\,$m in our training configuration.

\begin{table}[t]
    \centering
    \caption{Masking Ablation Results}
    \label{tab:masking_metrics}
    \footnotesize
    \setlength{\tabcolsep}{3pt}
    \renewcommand{\arraystretch}{1.12}
    \begin{tabular}{@{}ccc@{}}
        \toprule
        \textbf{Metric} & \textbf{w/o mask} & \textbf{Ours} \\
        \midrule
        Valid rollout (s) $\uparrow$ & $15.69\pm1.32$ & $19.46\pm0.52$ \\
        Lin. velocity error (m/s) $\downarrow$ & $0.194\pm0.003$ & $0.162\pm0.003$ \\
        Planar feet MPE (m) $\downarrow$ & $0.182\pm0.002$ & $0.167\pm0.002$ \\
        Wrist MPE (m) $\downarrow$ & $0.115\pm0.003$ & $0.079\pm0.004$ \\
        \bottomrule
    \end{tabular}
    \tabnote{Mean $\pm$ SD over 20 rollouts per variant, each capped at 20\,s.}
\end{table}

\subsection{Wrist-Equivalent Force Estimation and Contact-Load Adaptation}

\begin{figure*}[!t]
    \centering
    \includegraphics[width=0.72\textwidth]{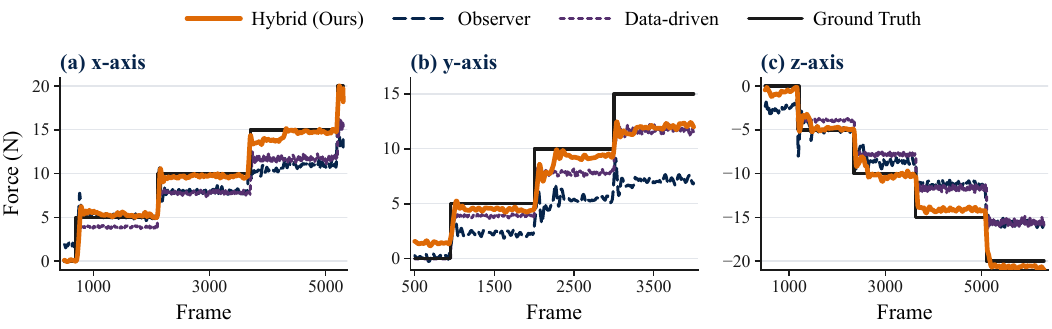}
    \caption{Representative single-trial real-world $x/y/z$ wrist-equivalent interaction-force estimates at staged $0,5,10,15,20\,$N loads. Each load is repeated 10 times with digitally synchronized force-gauge references.}
    \label{fig:force_real}
\end{figure*}

\begin{figure*}[!t]
    \centering
    \begin{subfigure}[b]{0.35\linewidth}
        \centering
        \includegraphics[width=\linewidth]{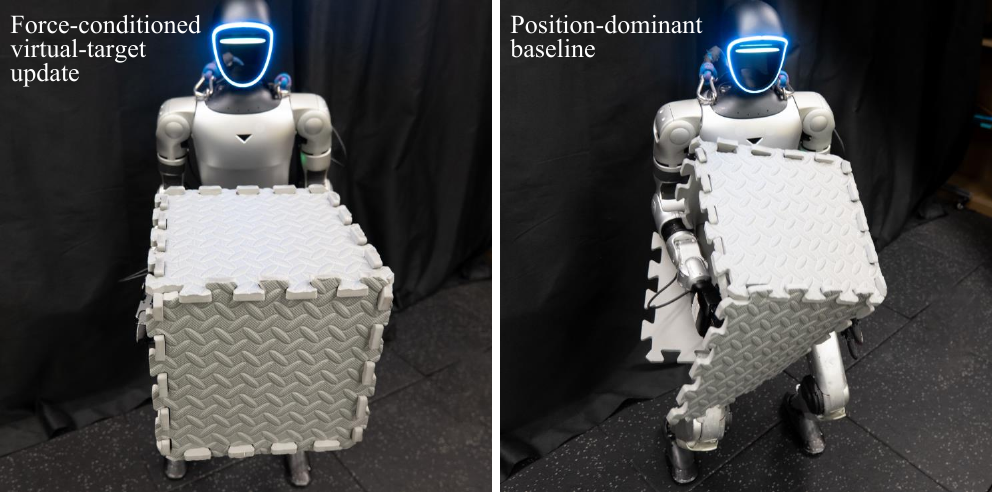}
        \caption{Foam-box lifting.}
        \label{fig:foam_box}
    \end{subfigure}\hfill
    \begin{subfigure}[b]{0.17\linewidth}
        \centering
        \includegraphics[width=\linewidth]{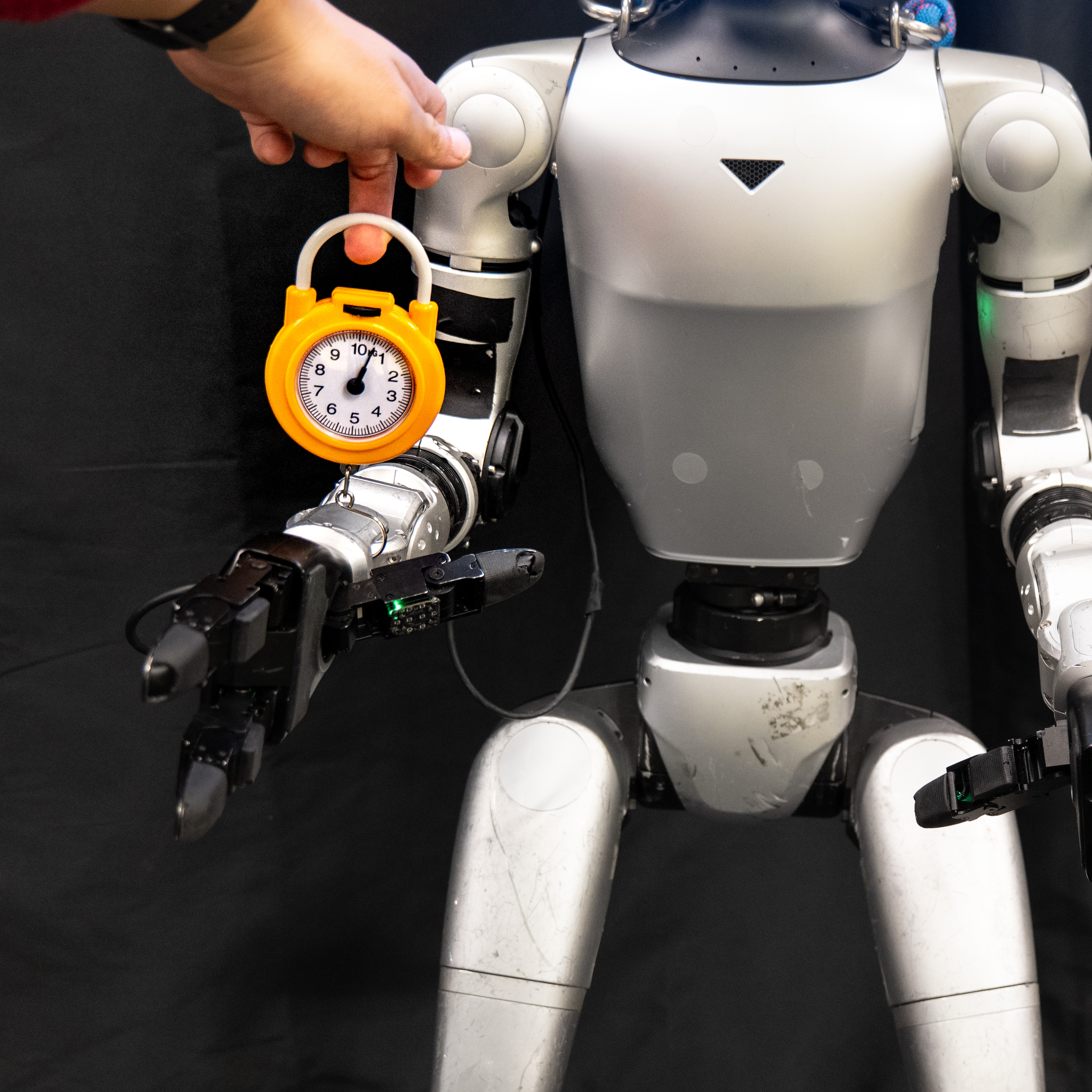}
        \caption{Force-test setup.}
        \label{fig:force_real_setup}
    \end{subfigure}\hfill
    \begin{subfigure}[b]{0.46\linewidth}
        \centering
        \includegraphics[width=\linewidth]{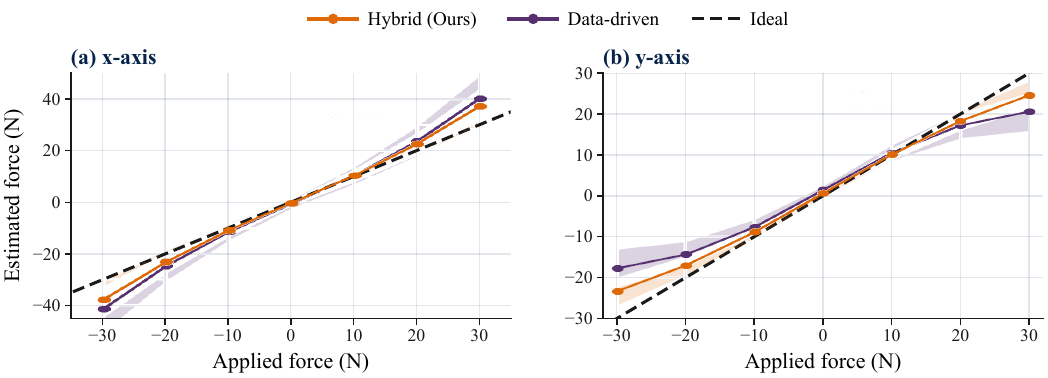}
        \caption{Planar $x/y$ loads.}
        \label{fig:force_xy}
    \end{subfigure}
    \caption{Contact-load adaptation and wrist-equivalent force estimation. (a) Foam-box lifting: feedback enabled (left) reduces visible deformation relative to feedback disabled (right). (b) Force-test setup. (c) Planar-load evaluation under external forces in $[-30,+30]\,$N: the hybrid estimator records lower error than the learned-only estimator. Shading denotes one standard deviation over ten repeats, using synchronized force-gauge references.}
    \label{fig:force_setup_xy}
\end{figure*}

We compare observer-only, learned-only, and hybrid force estimates under staged wrist loads (Fig.~\ref{fig:force_real}). Fig.~\ref{fig:force_setup_xy} presents the contact-adaptation test and planar-force evaluation. A fragile-box comparison switches the desired-force-conditioning pathway on or off while keeping the robot, object, nominal wrist trajectory, and task fixed. The contact-force column in Table~\ref{tab:baseline_comparison} separately compares complete controllers under identical wrist-position targets.

\begin{figure*}[!t]
    \centering
    \includegraphics[width=0.8\textwidth]{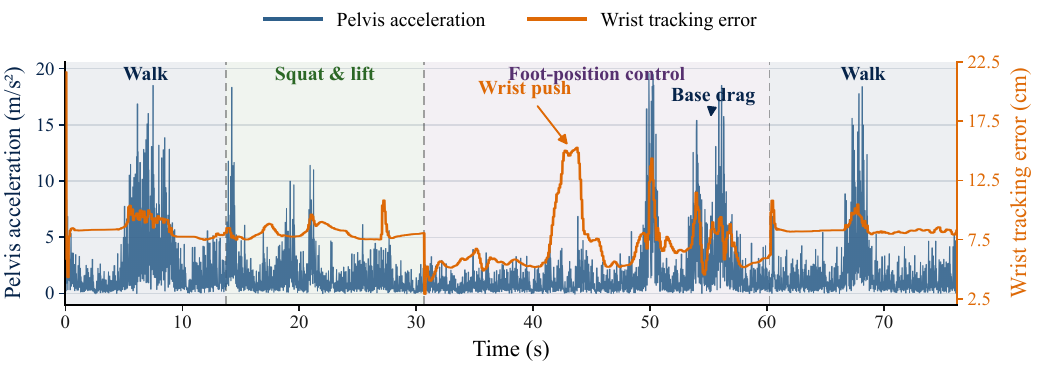}
    \caption{Walking, squat/lift, and foot-position control: wrist errors and pelvis acceleration briefly deviate and recover under wrist pushes and base drags.}
    \label{fig:pelvis_acc_wrist}
    \vspace{-3pt}
\end{figure*}


\noindent\textbf{Contact-sensitive teleoperation:}
We lift the same foam box with or without force feedback (Fig.~\ref{fig:force_setup_xy}\subref{fig:foam_box}), using identical policy weights, initial pose, nominal wrist trajectory, and lift target. Disabling feedback tracks nominal wrist targets directly; enabling it adapts them using estimated interaction forces. Both complete the lift, but feedback (left) produces less visible deformation than the baseline (right).

\noindent\textbf{Force estimation accuracy:}
We evaluate force estimation on G1 hardware using manually applied wrist loads measured with a handheld force gauge (Fig.~\ref{fig:force_setup_xy}\subref{fig:force_real_setup}). The gauge is aligned with each evaluated axis, and its readings provide a digitally logged, synchronized reference. The deployed estimator uses arm joint states and reconstructed PD torques without wrist force/torque sensors. Throughout estimator evaluation, $\mathbf f_{\mathrm{cmd}}=\mathbf 0$ and every gauge load is externally applied. We repeat each load condition 10 times for staged $0$--$20\,$N loads on all axes, signed planar forces in $[-30,+30]\,$N, and held-out vertical loads of 35/40/50\,N. Fig.~\ref{fig:force_real} shows representative trials; shading in Fig.~\ref{fig:force_setup_xy}\subref{fig:force_xy} shows one standard deviation over ten trials. The learned-only baseline follows UniFP's proprioceptive-history estimator design~\cite{unifp}.

The hybrid estimate follows the gauge more closely than observer-only and learned-only variants for staged loads (Fig.~\ref{fig:force_real}) and records lower planar error than learned-only estimation (Fig.~\ref{fig:force_setup_xy}\subref{fig:force_xy}). Table~\ref{tab:force_trace_error} quantifies all three estimators on representative staged-load traces. At held-out vertical loads of 35/40/50\,N, hybrid RMSEs are 8.28/9.66/12.35\,N.

\begin{table}[t]
    \centering
    \caption{Force-Estimation Errors under Staged Loads}
    \label{tab:force_trace_error}
    \footnotesize
    \setlength{\tabcolsep}{3.2pt}
    \renewcommand{\arraystretch}{1.12}
    \begin{tabular}{@{}lccc@{}}
        \toprule
        \textbf{Estimator} & \textbf{$x$ axis} & \textbf{$y$ axis} & \textbf{$z$ axis} \\
        \midrule
        Hybrid & $\mathbf{0.81/0.50}$ & $\mathbf{1.97/1.57}$ & $\mathbf{0.87/0.67}$ \\
        Observer-only & $2.83/2.29$ & $5.15/4.43$ & $3.02/2.58$ \\
        Learned-only & $2.43/2.19$ & $2.23/1.92$ & $2.88/2.47$ \\
        \bottomrule
    \end{tabular}
    \tabnote{Entries are RMSE/MAE in N for the traces in Fig.~\ref{fig:force_real}.}
\end{table}

\subsection{Hardware Tracking and Perturbation Recovery}

\noindent\textbf{Real-world tracking and perturbation recovery:}
Hardware tests cover walking, squat/lift, and commanded planar foot placement; Fig.~\ref{fig:pelvis_acc_wrist} reports wrist-tracking error and pelvis acceleration in representative runs. A separate dynamic-reaching evaluation yields 3D wrist-position and planar 2D foot-position RMSEs below 8\,cm and 10\,cm, respectively. Under the evaluated 30\,N wrist push and 50\,cm body drag, representative traces show transient errors followed by recovery with bounded pelvis acceleration.

\subsection{Demonstration Collection and Downstream Learning}

\noindent\textbf{Diffusion Policy as a validation:}
We use Diffusion Policy as an upstream visuomotor planner and \textbf{FORGE} as the only whole-body joint-action executor. One cross-payload model maps head-mounted RGB and task state to multi-granularity command chunks. Training pools 80 demonstrations from each of four payload conditions (320 total).

We evaluate this pipeline over 20 hardware trials for each payload on the squat-and-lift-box transport task with empty, 0.5\,kg, 1\,kg, and 2\,kg boxes. A trial succeeds only when the robot completes the prescribed transport sequence without task termination or human intervention. As shown in Table~\ref{tab:dp_box_lifting}, the complete pipeline succeeds in 20/20, 20/20, 18/20, and 18/20 trials, corresponding to success rates of 100\%, 100\%, 90\%, and 90\%, respectively. Assuming independent trials, the 95\% Wilson intervals are 83.9--100\% for 20/20 successes and 69.9--97.2\% for 18/20 successes. The same Diffusion Policy--\method pipeline completes the transport task across the tested 0--2\,kg payload range.

\begin{table}[t]
    \centering
    \caption{Cross-Payload Box-Transport Success Rates}
    \label{tab:dp_box_lifting}
    \footnotesize
    \setlength{\tabcolsep}{4pt}
    \renewcommand{\arraystretch}{1.12}
    \begin{tabular}{@{}lcccc@{}}
        \toprule
        \textbf{Payload} & \textbf{Empty} & \textbf{0.5\,kg} & \textbf{1\,kg} & \textbf{2\,kg} \\
        \midrule
        Success rate & 100\% & 100\% & 90\% & 90\% \\
        \bottomrule
    \end{tabular}
    \tabnote{20 hardware trials per payload. One cross-payload Diffusion Policy with \method is trained on 80 demonstrations per payload condition.}
\end{table}

\section{Conclusion}
\method combines multi-granularity commands, sensorless force estimation, and asymmetric residual whole-body control. Hardware experiments demonstrate task-space tracking, contact-aware manipulation, and execution of Diffusion Policy commands across tested payloads.

\FloatBarrier
\raggedbottom
\bibliographystyle{IEEEtran}
\bibliography{references}
\end{document}